\documentclass{svproc}
\usepackage{url}
\usepackage{cite}
\usepackage{amsmath,amssymb,amsfonts}
\usepackage{algpseudocode}

\usepackage{lipsum}
\usepackage{mwe}

\usepackage{amsmath}
\usepackage{soul}
\usepackage{subcaption}
\usepackage[table,xcdraw]{xcolor}
\usepackage[ruled,vlined]{algorithm2e}
\usepackage{graphicx}
\usepackage{textcomp}
\usepackage{xcolor}
\usepackage{soul}
\usepackage{booktabs}
\usepackage[font=small,skip=5pt]{caption}

\usepackage{verbatim}
\usepackage[scaled=1, varl]{inconsolata}

\usepackage{comment}

\usepackage{array}

\usepackage{listings}
\usepackage{xcolor}

\colorlet{punct}{red!60!black}
\definecolor{background}{HTML}{EEEEEE}
\definecolor{delim}{RGB}{20,105,176}
\colorlet{numb}{magenta!60!black}

\lstdefinelanguage{json}{
    basicstyle=\normalfont\ttfamily,
    stepnumber=1,
    numbersep=4pt,
    showstringspaces=false,
    breaklines=true,
    backgroundcolor=\color{background},
    literate=
     *{0}{{{\color{numb}0}}}{1}
      {1}{{{\color{numb}1}}}{1}
      {2}{{{\color{numb}2}}}{1}
      {3}{{{\color{numb}3}}}{1}
      {4}{{{\color{numb}4}}}{1}
      {5}{{{\color{numb}5}}}{1}
      {6}{{{\color{numb}6}}}{1}
      {7}{{{\color{numb}7}}}{1}
      {8}{{{\color{numb}8}}}{1}
      {9}{{{\color{numb}9}}}{1}
      {:}{{{\color{punct}{:}}}}{1}
      {,}{{{\color{punct}{,}}}}{1}
      {\{}{{{\color{delim}{\{}}}}{1}
      {\}}{{{\color{delim}{\}}}}}{1}
      {[}{{{\color{delim}{[}}}}{1}
      {]}{{{\color{delim}{]}}}}{1},
}

\makeatletter
\def\hlinewd#1{%
  \noalign{\ifnum0=`}\fi\hrule \@height #1 \futurelet
   \reserved@a\@xhline}
\makeatother

\def\BibTeX{{\rm B\kern-.05em{\sc i\kern-.025em b}\kern-.08em
    T\kern-.1667em\lower.7ex\hbox{E}\kern-.125emX}}

\begin{document}
\mainmatter              % start of a contribution
\title{Growing Artificial Neural Networks}
%
%\titlerunning{Hamiltonian Mechanics}  % abbreviated title (for running head)
%                                     also used for the TOC unless
%                                     \toctitle is used
%
\author{John Mixter \and Ali Akoglu}
\authorrunning{John Mixter and Ali Akoglu} % abbreviated author list (for running head)
%
%%%% list of authors for the TOC (use if author list has to be modified)
\tocauthor{John Mixter and Ali Akoglu }
\institute{University of Arizona, Tucson, AZ 85721, USA,\\
\email{\{jmixter6011,akoglu\}@email.arizona.edu}
%,\\ WWW home page:
%\texttt{http://users/\homedir iekeland/web/welcome.html}
%\and
%Universit\'{e} de Paris-Sud,
%Laboratoire d'Analyse Num\'{e}rique, B\^{a}timent %425,\\
%F-91405 Orsay Cedex, France
}

\maketitle              % typeset the title of the contribution

\begin{abstract}
Pruning is a legitimate method for reducing the size of a neural network to fit in low SWaP hardware, but the networks must be trained and pruned offline. We propose an algorithm, Artificial Neurogenesis (ANG), that grows rather than prunes the network and enables neural networks to be trained and executed in low SWaP embedded hardware. ANG accomplishes this by using the training data to determine critical connections between layers before the actual training takes place. Our experiments use a modified LeNet-5 as a baseline neural network that achieves a test accuracy of 98.74\% using a total of 61,160 weights. An ANG grown network achieves a test accuracy of 98.80\% with only 21,211 weights.  %When the ANG grown network is pruned, the size of the network drops to 18,167 weights while maintaining 98.80\% testing accuracy.
\keywords{neural network, pruning, dynamic growth}
\end{abstract}

%\vspace{-5mm}
\section{Introduction}
\label{sec:introduction}
The low size, weight and power (SWaP) requirements in embedded systems restrict the amount of memory and processing power available to execute a neural network.  The state-of-the-art neural networks tend to be quite large \cite{alexnet:1}\cite{Simonyan:Net} and fitting them into severely constrained hardware has proven to be difficult \cite{Zeiler:Net}\cite{Guo:FPGA}\cite{Omondi:FPGA}. The challenge of fitting a neural network into low SWaP hardware can be explained by examining neural network requirements - memory and computations.  A connection between two layers represents an input multiplied by a weight. The products of these calculations are then summed together to produce an input for the next layer. The weights, inputs and outputs require memory.  The larger the neural network, the more resources needed to support its execution. This poses as the barrier for creating neural networks that can be trained and executed in embedded systems. It has been known for decades that neural networks we design tend to be over-built \cite{LeCun:prune}\cite{Hassibi:prune}.  Studies have shown that in some cases more than 97\% of the weights and connections of a trained neural network can be eliminated without significant accuracy loss \cite{Han:Stanford}.  The process of eliminating unnecessary weights and connections is known as pruning. Because pruning can significantly reduce the size of a neural network, it is possible to execute inference in low SWaP hardware.  But, this does not enable training in hardware because a full-sized network must fit in hardware \textit{before} it can be pruned.
%In order to both train and execute inference in low SWaP embedded hardware a new method is needed.  
In that case growing a network from a small initial network would be the desired approach when restricted to low SWaP hardware  \cite{Macleod:grow} \cite{Ozan:grow}. We propose an algorithm, Artificial Neurogenesis (ANG), that allows growing small and accurate neural networks that are small enough to be both trained \textit{and} executed on-chip by 
%analyzing perceptron outputs to 
determining critical connections between layers. 

%within embedded hardware. The algorithm we call Artificial Neurogenesis 
%The ANG works by analyzing perceptron outputs to determine critical connections between layers. 
%Using ANG we are able to grow neural networks that are small enough to be both trained \textit{and} executed on-chip.

%The remainder of 
This paper is organized as follows. 
In Section \ref{sec:NN Pruning}, we discuss several methods for neural network size reduction and describe our baseline neural network. In Section \ref{sec:GrowingNN}, we introduce our method for growing a neural network as an alternative and present the ANG algorithm. We conduct parameter sweeps and experimentally determine the  final architecture in Section \ref{sec:NeurogenesisExperiments} followed by a detailed comparison with respect to the state-of-the-art pruning methods in Section \ref{sec:NeurogenesisAnalysis}.  Finally we present our conclusions and future work in Section \ref{sec:Conclusion}.
%\section{Motivation}
%DSSoC and design challenges.
%technology scaling.

%State of the art.

%Methods to hardware and configuration mapping. why existing solutions are not enough for DSSoC. Talk about there is no unified platform to validate the hardware, configuration, scheduler, and application concurrently.

%Exploring the hardware configuration, applciations,
%This is portable testbed where end user can run it on any platform. 

%\section{Literature survey}
\vspace{-2mm}
\section{Neural Network Pruning}
\vspace{-2mm}
\label{sec:NN Pruning}
%To the best of our knowledge, there is no network growing based approach in the literature. 

Pruning and growing based methods are fundamentally different in terms of their objectives. Growing based methods focus on creating optimal neural network architectures in terms of accuracy by iterative processes of adding entire nodes or layers and noting the results. They are not concerned about whether or not the final architectures are larger or smaller than manually designed architectures. The network grows as long as there is improvement in accuracy regardless of network size \cite{Macleod:grow}. Pruning based methods on the other hand start with a well-trained network and iteratively identify and eliminate connections that do not impact inference accuracy \cite{LeCun:prune}. Their goal is to reduce the network to the smallest size possible while maintaining the original accuracy.  

Similar to the pruning methods, we are concerned about the final size of the neural network. Therefore in our literature review we focus on results of pruning methods and compare our research against algorithms specifically designed to reduce the size of networks while maintaining good accuracy. For each method, we analyze the relationship between network accuracy and degree of pruning measured in terms of percentage of weights removed by referring to their reported results based on the MNIST data set.

The five prominent pruning algorithms we cover in our literature review for comparison all execute on the MNIST dataset.  Three of the architectures ~\cite{Babaeizadeh:Illinois} ~\cite{Han:Stanford} ~\cite{Srinivas:IIS} are based on LeNet-5, part of which we use as a seed network. The other two architectures ~\cite{Blundell:Google} ~\cite{Tu:ASU} are included for a more comprehensive analysis. 

%Pruning is the current approach for reducing the size of a trained neural network.% ~\cite{mit19}. 
%It eliminates weights and connections that do not contribute significantly to the network accuracy \cite{LeCun:prune, Hassibi:prune}. 
%In this section, we present a literature review of pruning methods. 

%The MNIST data set contains 70k images of handwritten digits.  There are ten classes in the data set one for each digit. The data set contains 10k testing and 60k training images.  
%The \emph{Bayes by Backprop} algorithm ~\cite{Blundell:Google} is a pruning method that is capable of learning neural network weight probability distribution and exploring the weight redundancies in the networks. 
Blundel et al.~\cite{Blundell:Google} 
introduce the \emph{Bayes by Backprop} algorithm for learning neural network weight probability distribution and exploring the weight redundancies in the networks. This algorithm is able to prune the weights that have a low signal to noise ratio and remove 95\% of the weights without significant accuracy reduction. Han et al.~\cite{Han:Stanford} 
propose a three-step process  to learn important connections in a neural network where they first train the network to learn which connections are important, then use regularization to increase and remove the number of weights that are near zero, and finally retrain the pruned network to maintain accuracy. This method is able to remove 91.7\% of weights and neurons for an accuracy of 99.23\%. 
%Blundel et al.~\cite{Blundell:Google} describe a way to learn the neural network weight probability distribution.  Their algorithm, \emph{Bayes by Backprop}, regularizes the weights by minimizing the compression cost. They explore the weight redundancies in the networks they trained using the \emph{Bayes by Backprop}.  This is accomplished by pruning the weights that have a low signal to noise ratio. Using this method they are able to remove 95\% of the weights without significant accuracy reduction.  
%Han et al.~\cite{Han:Stanford} address the problem of fitting neural networks in hardware by attempting to learn important connections in a neural network using a three-step process.  First, they train the network to learn which connections are important. Next, they use regularization to increase and remove the number of weights that are near zero. Finally, the network is retrained after the pruning to maintain accuracy. Any neurons found to have zero input or output connections are eliminated. Their method of pruning and retraining removes 91.7\% of weights and neurons for an accuracy of 99.23\%.
Srinivas and Babu~\cite{Srinivas:IIS} 
take a different approach to pruning neural networks. Their algorithm is designed to find sets of weights that are similar. The inputs associated with similar weights are added together and their sum is multiplied by the single weight value. In the case where there are no equal weights, they find weights that are close in value by calculating their 'saliency'.  %They create a 2D matrix of saliency values for all pairs of weights, choose the smallest entry in the matrix, add the associated inputs and multiply with the single weight value. 
Their pruning method substantially reduces the number of weights in the network, but suffers from a significant drop in test accuracy compared to~\cite{Han:Stanford}. 
Babaeizadeh et al.~\cite{Babaeizadeh:Illinois} 
propose a method that works on an entire neuron instead of its individual weights. Their approach relies on merging the neurons that demonstrate high neuron activation correlation. The pruning occurs during training and allows neurons that are not fully correlated to be merged and then retrained to compensate for any accuracy loss. %Using this method, they are able to reduce the number of neurons in the LeNet-5 from 512 to 3 without impacting the network accuracy. 
Starting with a well trained network, they are able to remove 97.75\% of the weights without accuracy loss.  
%Babaeizadeh et al.~\cite{Babaeizadeh:Illinois} propose a method for pruning that works on an entire neuron instead of its individual weights. Their approach relies on merging the neurons that demonstrate high neuron activation correlation. The pruning occurs during training and allows neurons that are not fully correlated to be merged and then retrained to compensate for any accuracy loss. %Using this method, they are able to reduce the number of neurons in the LeNet-5 from 512 to 3 without impacting the network accuracy. 
%Starting with a well trained network, they are able to remove 97.75\% of the weights without accuracy loss.  
Tu et al.~\cite{Tu:ASU} 
propose a deep neural network compression method that starts with removing the unnecessary parameters and then uses Fisher Information to further reduce the parameter count. As a last step, they utilize a non-uniform fixed point quantization to assign more bits to parameters with higher Fisher Information estimates. Their research of using information theory for pruning has resulted in reducing the weights by 94.72\%. 
%Tu et al.~\cite{Tu:ASU} propose a method of deep neural network compression using the Fisher Information metric. %The Fisher Information is estimated through stochastic optimization that keeps track of second order network information. 
%They start by removing the unnecessary parameters and then use Fisher Information to further reduce the parameter count.  As a last step, they utilize a non-uniform fixed point quantization to assign more bits to parameters with higher Fisher Information estimates. Their research of using information theory for pruning has resulted in reducing the weights by 94.72\%.  

\begin{table}[t]
\small
\caption{Baseline Neural Network with 4 Layers.}\label{tbl:BaselineArch}
\centering
\begin{tabular}{|c|c|c|c|r|r|}
\hline
%\rowcolor[HTML]{EFEFEF} 
Layer      & Filter & Kernel&Stride &Perceptron & Weight \\ \hline
2D Conv    & 6       & 7      & 2      & 864         & 300     \\ \hline
3D Conv    & 50      & 7      & 4      & 450         & 14,750   \\ \hline
Full       &         &        &        & 100         & 45,100   \\ \hline
Classifier &         &        &        & 10          & 1,010    \\ \hline
\multicolumn{4}{|r|}{Totals}           & 1,424       & 61,160    \\ \hline
\end{tabular}
\vspace{-6mm}
\end{table}

\iffalse
\begin{table}[t]
\caption{Baseline Weight Magnitude Pruning}\label{tbl:BaselinePrune}
\centering
%\small
\begin{tabular}{|c|c|c|c|}
\hline
%\rowcolor[HTML]{EFEFEF} 
\% Removed & Weights & Test Accuracy & Error \\ \hline
0.00\% & 61,160 & 98.74\% & 1.26\% \\ \hline
%4.69\% & 58,289 & 98.70\% & 1.30\% \\ \hline
9.16\% & 55,555 & 98.70\% & 1.30\% \\ \hline
%15.02\% & 51,974 & 98.73\% & 1.27\% \\ \hline
%20.66\% & 48,526 & 98.77\% & 1.23\% \\ \hline
%26.39\% & 45,020 & 98.70\% & 1.30\% \\ \hline
%28.97\% & 43,444 & 98.69\% & 1.31\% \\ \hline
%35.61\% & 39,378 & 98.71\% & 1.29\% \\ \hline
40.83\% & 36,186 & 98.72\% & 1.28\% \\ \hline
%44.51\% & 33,935 & 98.71\% & 1.29\% \\ \hline
50.18\% & 30,468 & 98.63\% & 1.37\% \\ \hline
%55.07\% & 27,482 & 98.52\% & 1.48\% \\ \hline
%59.87\% & 24,544 & 98.55\% & 1.45\% \\ \hline
%64.52\% & 21,698 & 98.44\% & 1.56\% \\ \hline
69.92\% & 18,395 & 92.96\% & 7.04\% \\ \hline
%74.82\% & 15,403 & 69.24\% & 30.76\% \\ \hline
\end{tabular}

\end{table}
\fi

As a motivation for our network growing approach we also ran a pruning experiment. Our aim is to first demonstrate that using magnitude pruning method, we can reduce our baseline network to similar size of the more sophisticated pruning methods while achieving competitive accuracy. This will also serve as a comparison basis later when we introduce our growing method. Because both of our pruning and growing methods target the same fully connected layer, as a second aim, we will be in a position to demonstrate that we can grow a network that is smaller than a similar pruned network. The architecture we chose for our baseline used two sequential convolution layers followed by a fully connected layer as shown in Table \ref{tbl:BaselineArch}. The network has four layers, 61,160 32-bit weights and 221,950 connections. The first layer has six two-dimensional convolution filters each with a 7x7 kernel and a stride of two. The second layer is a three-dimensional convolution layer that has 50 filters each with a 7x7x6 kernel and a stride of four. The third layer is a fully connected layer with 100 perceptrons (100 FC). The fourth and final layer is the classifier, which has one perceptron for each class. Each of the ten perceptrons in the classifier layer is fully connected to the third layer. LeNet-5~\cite{LeNet:5} uses a similar architecture but with max pooling layers after each convolution. Max pooling layers help to reduce the number of weights but increase the required calculations. After the addition of a fully connected and classification layer, our baseline network has 0.5\% more weights than LeNet-5 but, requires 35\% fewer calculations to perform inference. To calculate size and number of connections, we use %Equations 1 and 2 for convolution layers.
Equations \ref{eq:w1} and \ref{eq:w2} for convolution layers. 

\begin{comment}
\footnotesize{
\begin{flushright}
Weights $=$ ({$Kernel_{Row}$} $\times$ {$Kernel_{Column}$} + Bias) $\times$ {$Filter_{Count}$}  $\ \ \ \ \ \ \   (1)$
\end{flushright}
}

\footnotesize{
Connections $=$ ({$Kernel_{Row}$} $\times$ {$Kernel_{Column}$} $+$ Bias) $\times$}  
\begin{flushright}
     $(\frac{Input Window Row - Kernel - Stride}{Stride})^2$ $\ \ \ \ \ \ \   (2)$
  \end{flushright}  
\end{comment}

\begin{equation}
    \label{eq:w1}
    \footnotesize
    Weights = ({Kernel_{Row}} \times {Kernel_{Column}} + Bias) 
    \times {Filter_{Count}}  
\end{equation}
\begin{multline}
    \label{eq:w2}
    \footnotesize
    Connections = ({Kernel_{Row}} \times {Kernel_{Column}} + Bias) \\ 
    \times (\frac{Input Window Row - Kernel - Stride}{Stride})^2 
\end{multline}

where ${Kernel_{Row}}$ is the number of rows and ${Kernel_{Column}}$ is the number of columns in the convolution filter.

To calculate size and number of connections for fully connected 
layers, we use Equations \ref{eq:w3} and \ref{eq:w4}.
%To calculate size and number of connections for fully connected layers we use:
%\begin{equation}
%{\scriptstyle
%Weights = Number of Perceptrons \times (Previous Layer Output Count + Bias)  }
%\end{equation}

\begin{equation}
    \label{eq:w3}
    \footnotesize
Weights = \#ofPerceptrons \times ({PreviousLayer_{Output Count}} + Bias) 
\end{equation}
\begin{equation}
    \label{eq:w4}
    \footnotesize
Connections = Weights - \#ofPerceptrons
\end{equation}

Weight magnitude pruning is used to reduce the size of our baseline network. 
%%The perceptron output is the sum of the weight and input products. 
Weights close to zero have a small impact on the output sum. A threshold is used to determine how far away from zero a weight value has to be before it is removed. 
%%We are targeting the fully connected layers where each connection is associated with a unique weight.  Removing a weight has the effect of removing a connection.  
%In Table \ref{tbl:BaselinePrune} we show test accuracy with respect to percentage of weights removed. 
With this method we can remove 50\% of weights without sacrificing the accuracy. In Table \ref{tbl:PruneCompare} we compare various pruning methods based on the percentage of weights removed and the change in accuracy with respect to the original network before pruning. The method of Babaeizadeh et al.~\cite{Babaeizadeh:Illinois} 
is able to prune down to the smallest network without accuracy loss.  Pruning approach of Han et al.~\cite{Han:Stanford} 
has the next most notable performance in terms of network size reduction with a slightly improved accuracy. Our baseline network has the least number of weights pruned but still resulted in the second smallest network and third smallest loss in accuracy.

\begin{table}[t]
\small
\caption{Comparing Pruned Networks}\label{tbl:PruneCompare}
\centering
\begin{tabular}{|c|c|c|c|}
\hline
%\rowcolor[HTML]{EFEFEF} 
Neural Network & Removed & Weights & $\Delta$ Accuracy \\ \hline
Babaeizadeh et al. (LeNet) & 97.8\% & 13,579 & 0.00\% \\ \hline
Tu et al. (FC) & 94.7\% & 31,972 & -0.92\% \\ \hline
Han et al. (LeNet) & 91.7\% & 36,000 & 0.03\% \\ \hline
Blundell et al. (FC) & 98.0\% & 48,000 & -0.15\% \\ \hline
Srinivas et al. (LeNet) & 83.5\% & 71,000 & -0.76\% \\ \hlinewd{1pt}
Pruned Baseline (LeNet) & 50.0\% & 30,468 & -0.11\% \\ \hline
\end{tabular}
\vspace{-5mm}
\end{table}
%Our analysis shows that we can use pruning to reduce our baseline accuracy to a size that can fit into low SWaP hardware.  Considering that all other networks start with a weight count above 400,000 before pruning, they are not suitable for fitting in embedded hardware and they all have to be trained and pruned offline. 

In summary, the methods mentioned above %are all capable of dramatically reducing 
prune large number of connections and reduce the size of networks without sacrificing accuracy.  %Because 
%Pruning large numbers of connections does not significantly degrade the accuracy. 
This indicates many weights in the fully connected layer are unnecessary, which leads us to a question: as we forward-propagate training data through a neural network, can the outputs of layers be analyzed to find only necessary (critical) connections? If this is possible then the critical question is: \textit{Can we find an optimal architecture by analyzing the output of perceptrons before training?} If so, this would eliminate the need to train and then prune an entire neural network offline before implementing in hardware. We introduce our approach which answers these questions in the next section.

%\section{Programming Model}
%\label{sec:prog_model}

%\section{Scheduler Architecture}
%\input{input_sections/scheduler_architecture.tex}
\vspace{-2mm}
\section{Artificial Neurogenesis}
\vspace{-2mm}
\label{sec:GrowingNN}
\subsection{Background}
\vspace{-2mm}
%To avoid confusion when explaining connections between layers, we will refer to the layer that produces the output values as the \textit{source layer} and the layer that receives input values as the \textit{destination layer}. 

For reference, when explaining connections between layers, we will refer to the layer that produces output values as the \textit{source layer} and the layer receiving values as the \textit{destination layer}.  We treat \textit{perceptrons} as atomic units whose inputs are connected to the outputs of perceptron(s) in the \textit{source layer}.  Our perceptron is a simple multiply and accumulate engine whose sum is applied to a non-linear function, which is then passed on to perceptron inputs in the next layer. Artificial Neurogenesis (ANG), our method of growing a neural network, is employed where a fully connected layer would normally reside.  It grows the neural network by adding perceptrons to a new \textit{destination layer} while making only critical connections from the outputs in the \textit{source layer}.  %The key to ANG is determining the number of perceptrons needed in the new \textit{destination layer} and finding the critical outputs in the \textit{source layer}.

%To help orient the reader we need to define a perceptron as it relates to our algorithm.  

A \textit{Seed Network} is the starting point for ANG. The last layer of a \textit{Seed Network} is the \textit{source layer} for the ANG algorithm.  Because a \textit{Seed Network} determines the minimum size of the neural network, it needs to be as small as possible. When working with two-dimensional input data, convolution layers are a good choice for the \textit{Seed Network}.  A convolution layer requires very few weights and the feature maps they produce are the same size or smaller than their input data. Smaller feature maps require fewer connections to the \textit{destination layer}.

Critical connections between \textit{source} and \textit{destination layers} are determined by analyzing the perceptron outputs of the \textit{source layer} as we forward-propagate training data through the \textit{Seed Network}. The \textit{source layers} produce feature maps whose values change significantly as data from different classes are presented to the \textit{Seed Network} input.  We use these feature map differences to determine which \textit{source layer} perceptrons are outputting critical data.

\begin{algorithm}[t]
\caption{Extreme Class Member Search}
\label{algo:SortClassMembers}
\SetAlgoLined
\KwResult{Class Members Sorted by Mean Squared Error Between Source Layer Output and Source Layer Average Output}

    Create a \textit{Seed Network}\;
    Attach a Temporary Classifier\;
    Prime \textit{Seed Network}\;
    Remove Temporary Classifier\;

 \For{Each Class in Data Set}
 {
    Count \begin{math}\gets\end{math} Number of Class Members\;
    \For{Each Perceptron in Source Layer}
    {   
        Sum[\textit{Perceptron}] \begin{math}\gets\end{math} 0\;
    }

    \For{Each Member in Class}
    {  
        Forward-Propagate Class Member\;
        \For{Each Perceptron in Source Layer}
        {   
            Sum[\textit{Perceptron}] {+}= \textit{Perceptron Output}\;
        }
    }
    \For{Each Perceptron in Source Layer}
    {   
        Avg[\textit{Perceptron}] \begin{math}\gets\end{math} Sum[\textit{Perceptron}] $\div$ Count\;
    }
    \For{Each Member in Class}
    {  
        Error[\textit{Member}] \begin{math}\gets\end{math} 0\;
        Forward-Propagate Class Member\;
        \For{Each Perceptron Output in Source Layer}
        {   
            Error[\textit{Member}] {+}= Error( Avg[\textit{Perceptron}] - \textit{Output} )\;
        }
    }
    Sort Members in Class by Error[\textit{Member}]\;
 }
\end{algorithm}

\begin{algorithm}[t]
\SetAlgoLined
\KwResult{New Destination Layer Connected to Source Layer Critical Outputs}
 \For{Each Class in Data Set}
 {
    \For{First and Last Members in Class}
    {  
        Forward-propagate Class Member\;

        Sum \begin{math}\gets\end{math} 0\;
        \For{Each Source Layer Perceptron Output}
        {   
             Sum {+}= \textit{Output}\;
        }
        Average \begin{math}\gets\end{math} Calculate Output Mean\;
        Sum \begin{math}\gets\end{math} 0\;
        \For{Each Source Layer Perceptron Output}
        {   
            Sum {+}= (Average - \textit{Output})\begin{math}^2\end{math}\;
        }
        \begin{math}\sigma\end{math} \begin{math}\gets\end{math} Calculate Standard Deviation\;
        Add Extreme Perceptron to Destination Layer\;
        \For{Each Source Layer Perceptron Output}
        {   
            \eIf{\textit{Output} is \begin{math} \pm x\sigma\end{math} from Mean}
            {
                Connect Perceptron Input to \textit{Output}\;
            }
            {
                This is Not a Critical Connection\;
            }
        }
    }
 }
 \caption{Find and Connect Critical Outputs}
  \label{algo:CriticalOutputs}
\end{algorithm}

\begin{algorithm}[t]
\caption{Artificial Neurogenesis}
\label{algo:PrimeTrain}
\SetAlgoLined
    Create Seed Network (Table~\ref{tbl:SeedNetwork})\;
    Prime Seed Network\;
    Remove Seed Network Classifier\;
    Add a Destination layer\;
    Add Classifying layer\;
    \While{Accuracy not Achieved}
    {
    Extreme Member Search (Algorithm~\ref{algo:SortClassMembers})\;
 
        \For{Each Extreme Member}
        {
    Find and connect critical outputs (Algorithm~\ref{algo:CriticalOutputs})\;
     Connect Extreme Perceptron output to Classifier\;
  }
  Train network using ALL training data\;
  Remove found extreme members from training set\;
  }

\end{algorithm}
\vspace{-2mm}
\subsection{Artificial Neurogenesis Algorithms}\label{sec:neurgenesis}
\vspace{-2mm}
ANG consists of two algorithms designed to find the critical connections between layers.  %Each algorithm analyzes the outputs of the \textit{source layer} perceptrons.  
Algorithm \ref{algo:SortClassMembers} searches for the two most extreme members of each class.  For every class, \textit{source layer} output averages are calculated as each member is forward-propagated through the \textit{Seed Network}.  The members of a class that produce outputs most and least similar to the average outputs are chosen as extreme members. Algorithm \ref{algo:CriticalOutputs} uses the extreme members of each class to determine the critical \textit{source layer}  perceptron outputs, which are then connected to the \textit{destination layer} perceptron inputs.

As a preprocessing step, before executing Algorithms 1 and 2, we first build a \textit{Seed Network} to which a temporary classifier is connected.  Once the temporary classifier is fully connected, the \textit{Seed Network} is \textit{primed} by training on all available training data. %, in this case 57,000 images.  
A \textit{priming cycle} occurs when all the training data has been forward and  back-propagated once.  After several \textit{priming cycles}, the temporary \textit{Seed Network} classifier is removed and the outputs of the \textit{source layer} perceptrons are ready to be analyzed by Algorithms \ref{algo:SortClassMembers} and \ref{algo:CriticalOutputs}.
\vspace{-2mm}
\subsubsection{Algorithm 1 - Extreme Member Search:$\ $ }
This algorithm is designed to find two members of each class that cause extreme feature maps to be generated at the \textit{source layer} outputs. %To find the class members that create extreme feature maps, we first need to find the average feature map for each class.
As shown in Algorithm \ref{algo:SortClassMembers}, one class at a time, all the members of the class are forward-propagated through the \textit{Seed Network}. After each class member is forward-propagated to the \textit{source layer}, a sum for each perceptron output is calculated.  After all of the class members have been forward-propagated, the average output value for each \textit{source layer} perceptron is determined by dividing the perceptron sums by the number of class members.  Together, all of the output averages form an average feature map for each class, and each member of the class is
%Once the average feature map for the class has been determined, each member of the class is 
forward-propagated through the \textit{Seed Network} again.  The feature map created by the \textit{source layer} for each class member is compared to the average feature map and the mean squared error between the two is calculated.  Members of the class are then sorted by their mean squared error.  After sorting, the first member of the class has a feature map that is most similar to the average and the last member has a feature map that is least similar to the average.  The first and last in the sorted class list represent the extreme members of that class.

\subsubsection{Algorithm 2 - Critical Connection Search:$\ $}
 In this phase we grow a \textit{destination} layer by adding and connecting new perceptrons to the \textit{source layer}.  We refer to the new perceptrons as \textit{extreme perceptrons}. For every class, we present the first member to the \textit{Seed Network} and forward-propagate.  The \textit{source layer} outputs are analyzed to calculate the average and standard deviation ($\sigma$) for all perceptron outputs.  Perceptrons in the \textit{source layer} whose outputs are \begin{math} \pm x\sigma\end{math} from the average are critical outputs, where \begin{math}  x\end{math} is a scaling factor.  These outputs are connected to the \textit{destination layer's} extreme perceptron.  After the extreme perceptron is connected, another extreme perceptron is added to the \textit{destination layer} and the last member of the class is presented to the \textit{Seed Network} input.  The \textit{Seed Network} is forward-propagated and the process is repeated.  After two extreme perceptrons for the class are connected, first member of the next class is presented to the \textit{Seed Network} and the whole process is repeated.  Algorithm \ref{algo:CriticalOutputs} shows the process for finding and connecting critical \textit{source layer} outputs.

Once two extreme perceptrons for each class have been added and connected to the \textit{source layer}, a classifying layer is connected to the new \textit{destination layer} and the entire network is trained. If the grown network does not meet accuracy goals, the newly grown \textit{destination layer} is augmented with additional extreme perceptrons.  The classes that do not perform well are the sources of input data to initiate further ANG cycles.  For each class that needs improvement, the second and second to last members are used as inputs just like the first and last were originally used.
\vspace{-2mm}
\subsubsection{Algorithm 3 - ANG Overview:$\ $}
As shown in Algorithm~\ref{algo:PrimeTrain}, the ANG is implemented based on the interactions between the extreme member search and critical connection search  algorithms. The algorithms are nested in two loops. The outside loop, which executes Algorithm \ref{algo:SortClassMembers}, is continued until the required accuracy is met. The inner loop executes Algorithm \ref{algo:CriticalOutputs} for each critical member in every class. It is important to note that after the inner loop finishes, a new set of extreme members are found in the outer loop. The extreme members that were just used to find critical connects are removed from consideration in the next iteration. This forces new critical connections to be found.
\vspace{-2mm}
\section{Artificial Neurogenesis Experiments}
\vspace{-2mm}
\label{sec:NeurogenesisExperiments}
In our experiments, we use the MNIST data set that contains 60,000 training and 10,000 testing images. We divided the training data set into 57,000 images for training and 3,000 images for validation and network tuning. The testing data held aside and is never used for tuning the networks.  

The training of our neural networks is divided into cycles.  Like a \textit{priming cycle}, a single training cycle is completed when all of the training data has been forward and back-propagated once.  At the end of every training cycle, we execute a validation inference and then randomize the training data.  The accuracy of the validation inference is compared to a current maximum.  If the validation inference accuracy is greater than the current maximum, the current maximum is set to the validation accuracy and testing inference is executed and the testing accuracy is noted.

Because experimental results are affected by the order in which training images are presented to the neural network, we have turned off random seeding so that the software generates identical sets of random images for each test. We do the same for weight initialization.  This helps to ensure any accuracy change is due solely to modifications we make to the network.

In each experiment, training is halted when one of the following \textit{stopping criteria} are met; the number of training cycles reaches 30, the validation accuracy reaches 100\%, or the validation accuracy does not increase over 20 consecutive training cycles. 

Our ANG experiments require us to choose a \textit{Seed Network}.  To ensure our experiments can be easily compared to our baseline (Table~\ref{tbl:BaselineArch}), it makes sense to use the same two convolution layers that our baseline network uses. For each experiment we prepare the \textit{Seed Network} as outlined in Section \ref{sec:neurgenesis}.  First, we build a \textit{Seed Network} (Table \ref{tbl:SeedNetwork}) to which we connect a temporary classifier.  We then randomize all weights and proceed to prime the network.  After priming, the \textit{Seed Network} is ready for the experiments. Before comparing networks grown with ANG to our baseline network, we conduct experiments for hyper-parameter selection that we cover in the following subsections.

\begin{table}[t]
\small
\caption{Seed Network plus Classifier}\label{tbl:SeedNetwork}
\centering
\begin{tabular}{|c|c|c|c|r|r|}
\hline
%\rowcolor[HTML]{EFEFEF} 
Layer      & Filter & Kernel & Stride & Perceptron & Weight \\ \hline
2D Conv    & 6       & 7      & 2      & 864         & 300     \\ \hline
3D Conv    & 50      & 7      & 4      & 450         & 14,750   \\ \hline
Classifier &         &        &        & 10          & 1,010    \\ \hline
\multicolumn{4}{|r|}{Totals}            & 1,324        & 16,060    \\ \hline
\end{tabular}
\vspace{-5mm}
\end{table}

\iffalse
\begin{enumerate}
\item Finding the minimum number of priming cycles needed to reach the maximum \textit{Seed Network} accuracy.
\item Finding the fraction of sigma that results in the maximum network accuracy based on the minimum priming cycle count.
\item Finding the best network architecture by sweeping the number of priming cycles looking for the highest accuracy based on the fraction of sigma derived.
\end{enumerate}

After identifying the best ANG hyper-parameters, we will present the following detailed experiments: 
\begin{enumerate}
\item Compare a fully connected network with the same perceptron count to our ANG grown network.
\item Prune the fully connected layer to the same weight count as the grown network and compare.
\item Investigate the grown layer sensitivity to pruning.
\item Investigate the classifying \textit{and} grown layer sensitivity to pruning.
\item Investigate the convolution layers sensitivity to pruning.
\item Determine the entire network sensitivity when pruning all layers.
\end{enumerate}
\fi
\vspace{-2mm}
\subsection{Finding Minimum Priming Cycle Count}

%\subsubsection{Purpose}
We prime the \textit{Seed Network} to move the weights out of a random state.  As the accuracy of the \textit{Seed Network} increases, the feature maps created by the \textit{source layer} become more focused and we are better able to find the critical outputs. As the priming cycles progress, accuracy saturation is expected. Therefore, our aim in this experiment is to determine the minimum number of priming cycles before saturation occurs.
%\subsubsection{Procedure}
%The \textit{Seed Network} is prepared as described earlier in this section.  

We prime the network allowing the validation inference to dictate when a testing inference is executed. The experiment ends when one of the stopping criteria are met. In Table \ref{tbl:SeedNetworkPriming}, we show Training cycle (57,000 images per cycle), Training accuracy, Validation accuracy on 3,000 images after training and Test accuracy on 10,000 images in the test data set. We observed saturation in \textit{validation accuracy} starting at 18 priming cycles through 30 cycles. Therefore we will use cycle 18 as the stopping point in the following experiments. %We also note that, although this is a small network, it is able to achieve 98.2\% testing accuracy.

%\subsubsection{Analysis}
%Table \ref{tbl:SeedNetworkPriming} shows the results of this experiment. The columns indicate the training cycle (57,000 images per cycle), the training accuracy on 57,000 images, validation accuracy on 3,000 images after training and the Test accuracy on the 10,000 images in the test data set.   We observe the saturation in validation accuracy at 18 priming cycles, which we will use as the stopping point in the following experiments. We also note that, although this is a small network, it is able to achieve 98.2\% testing accuracy.

\begin{table}[t]
\small
\caption{Seed Network Priming Accuracy}\label{tbl:SeedNetworkPriming}
\centering
\begin{tabular}{|c|c|c|c|}
\hline
%\rowcolor[HTML]{EFEFEF} 
Cycle & Train & Validate & Test \\ \hline
0 & 0.00\% & 0.00\% & 9.90\% \\ \hline
1 & 89.87\% & 96.18\% & 95.13\% \\ \hline
2 & 95.97\% & 97.94\% & 96.93\% \\ \hline
3 & 96.88\% & 98.22\% & 97.47\% \\ \hline
4 & 97.31\% & 98.33\% & 97.52\% \\ \hline
7 & 97.79\% & 98.63\% & 97.86\% \\ \hline
12 & 98.07\% & 98.83\% & 98.03\% \\ \hline
15 & 98.23\% & 98.84\% & 98.04\% \\ \hline
\textbf{18} & \textbf{98.32\%} & \textbf{98.86\%} & \textbf{98.20\%}\\ \hline
30 & 98.32\% & 98.86\% & 98.20\% \\ \hline
\end{tabular}
\vspace{-2mm}
\end{table}

\begin{table}[t]
\centering
\caption{Seed Network Scaling Factor Sweep vs. Number of Critical Connections and Test Accuracy when Validation Accuracy peaks}\label{tbl:SeedNetworkSigma}
\small
\begin{tabular}{|c|c|c|c|c|c|}
\hline
%\rowcolor[HTML]{EFEFEF} 
Scale & Critical  & Test  & Scale & Critical  & Test  \\ 
%\rowcolor[HTML]{EFEFEF}
Factor & Connect & Accy & Factor &Connect &Accy
\\
\hline
0.1 & 16,180 & 98.24\% & 0.9 & 20,876 & 98.38\% \\ \hline
0.2 & 17,063 & 98.49\% &\textbf{ 1.0} & \textbf{21,211} & \textbf{98.80\%} \\ \hline
0.3 & 17,832 & 98.36\% & 1.01 & 21,239 & 98.82\% \\ \hline
0.4 & 18,502 & 98.52\% & 1.1 & 21,472 & 98.62\% \\ \hline
0.5 & 19,133 & 98.59\% & 1.2 & 21,743 & 98.45\% \\ \hline
0.6 & 19,666 & 98.60\% & 1.3 & 22,009 & 98.56\% \\ \hline
0.7 & 20,135 & 98.44\% & 1.4 & 22,267 & 98.57\% \\ \hline
0.8 & 20,542 & 98.75\% & 1.5 & 22,529 & 98.71\% \\ \hline

%0.1 & 16,180 & 98.24\% &\textbf{ 1.0} & \textbf{21,211} & \textbf{98.80\%} \\ \hline
%0.2 & 17,063 & 98.49\% &\textbf{ 1.0} & \textbf{21,211} & \textbf{98.80\%} \\ \hline
%0.3 & 17,832 & 98.36\% & 1.1 & 21,472 & 98.62\% \\ \hline
%0.4 & 18,502 & 98.52\% & 1.2 & 21,743 & 98.45\% \\ \hline
%0.5 & 19,133 & 98.59\% & 1.3 & 22,009 & 98.56\% \\ \hline
%0.6 & 19,666 & 98.60\% & 1.4 & 22,267 & 98.57\% \\ \hline
%0.7 & 20,135 & 98.44\% & 1.5 & 22,529 & 98.71\% \\ \hline
%0.8 & 20,542 & 98.75\% & 
%0.9 & 20,876 & 98.38\% &  &  &  \\ \hline

\end{tabular}
\vspace{-5mm}
\end{table}

%\iffalse

%akoglu
\begin{comment}
\begin{figure}[t!]
\centering
    \begin{subfigure}[t]{0.1\textwidth}
        \centering
        \includegraphics[scale=0.15]{input_sections/images/A_Connect.jpg}
        \caption{}
        \label{sub:A}
    \end{subfigure}%
  ~  
     \begin{subfigure}[t]{0.1\textwidth}
        \centering
        \includegraphics[scale=0.15]{input_sections/images/B_Connect.jpg}
        \caption{}
        \label{sub:B}
    \end{subfigure}%
   ~ 
      \begin{subfigure}[t]{0.1\textwidth}
        \centering
        \includegraphics[scale=0.15]{input_sections/images/C_Connect.jpg}
        \caption{}
        \label{sub:C}
    \end{subfigure}%  
   ~
\begin{subfigure}[t]{0.1\textwidth}
        \centering
        \includegraphics[scale=0.15]{input_sections/images/D_Connect.jpg}
        \caption{}
        \label{sub:D}
    \end{subfigure}%  

\caption{Example Seed Network Outputs Based on Class 5. (a) No priming,(b) No Priming $\pm$ 1$\sigma$, (c) 18 Priming Cycles, (d)18 Priming Cycles$\pm$ 1$\sigma$, Black = No Connection
}
\label{fig:Connections}
\vspace{-5mm}
\end{figure}
%\fi
\end{comment}

\subsection{Finding the Best Scaling Factor $x$}\label{sub:scaling factor}
\vspace{-2mm}
%\subsubsection{Purpose}
When the extreme perceptrons are being connected to the \textit{source layer}, the $\sigma$ value determines how far from the average an output can be and still be considered critical.  In this experiment, using the minimum priming cycle count, our aim is to determine the value for the scaling factor $x$ that produces the most accurate network.
%\subsubsection{Procedure}

For each scaling factor shown in Table \ref{tbl:SeedNetworkSigma}, we prepare a \textit{Seed Network}, prime it to the minimum priming cycle count and analyze the \textit{source layer} output to determine $\sigma$.  We sweep the scaling factor value ranging from 0.1 to 1.5 in increments of 0.1 and apply over the $\sigma$ value to change the number of critical connections made. After the critical connections are made, a classifier is connected to the new \textit{destination layer} and the entire network is trained.
%\subsubsection{Analysis}
As shown in Table \ref{tbl:SeedNetworkSigma}, sweeping the scaling factor has the desired effect of changing the number of critical connections. We observe that both the test accuracy and the \textit{validation accuracy} peak at 1.01.  However, because the difference between this accuracy and the accuracy at a scaling factor of 1.0 is only 0.02\%, we choose the scaling factor of 1.0 to eliminate the need for a scaling hyper-parameter.

\color{black}
\begin{table}[t]
\caption{Priming Cycles vs. Critical Connections at peak Validation Accuracy (1$\sigma$)} \label{tbl:FullNetworkSigma}
\centering
\small
\begin{tabular}{|c|c|c|c|c|c|}
\hline
Cycle & Critical  & Test  & Cycle & Critical  & Test  \\ 
 & Connect & Accy &  &Connect &Accy \\ \hline
0 & 21,362 & 98.36\% & 13 & 21,100 & 98.77\% \\ \hline
1 & 21,731 & 98.28\% & 14 & 21,100 & 98.56\% \\ \hline
2 & 21,504 & 98.51\% & 15 & 21,069 & 98.72\% \\ \hline
3 & 21,370 & 98.56\% & 16 & 21,069 & 98.66\% \\ \hline
4 & 21,277 & 98.57\% & 17 & 21,069 & 98.56\% \\ \hline
5 & 21,277 & 98.64\% & 18 & 20,998 & 98.51\% \\ \hline
6 & 21,277 & 98.63\% & 19 & 20,998 & 98.45\% \\ \hline
7 & 21,211 & 98.44\% & 20 & 20,998 & 98.66\% \\ \hline
8 & 21,211 & 98.70\% & 21 & 20,953 & 98.65\% \\ \hline
9 & 21,211 & 98.58\% & 22 & 20,953 & 98.61\% \\ \hline
10& 21,211 & 98.64\% & 23 & 20,953 & 98.52\% \\ \hline
\textbf{11} & \textbf{21,211} & \textbf{98.80\%} & 24 & 20,953 & 98.75\% \\ \hline
12 & 21,100 & 98.69\% & 25 & 20,953 & 98.46\% \\ \hline

%0 & 21,362 & 98.36\% & 14 & 21,100 & 98.56\% \\ \hline
%2 & 21,504 & 98.51\% & 16 & 21,069 & 98.66\% \\ \hline
%4 & 21,277 & 98.57\% & 18 & 20,998 & 98.51\% \\ \hline
%6 & 21,277 & 98.63\% & 20 & 20,998 & 98.66\% \\ \hline
%8 & 21,211 & 98.70\% & 22 & 20,953 & 98.61\% \\ \hline
%10& 21,211 & 98.64\% & 23 & 20,953 & 98.52\% \\ \hline
%\textbf{11} & \textbf{21,211} & \textbf{98.80\%} & 24 & 20,953 & 98.75\% \\ \hline
%12 & 21,100 & 98.69\% & 25 & 20,953 & 98.46\% \\ \hline
\end{tabular}
\vspace{-4mm}
\end{table}

%\iffalse
%akoglu
\begin{comment}
Figure \ref{fig:Connections} is a graphical representation of the \textit{Seed Network} \textit{source layer} output values.  The four images are 50x9 feature maps produced when the first member of class 5, which is the extreme member most similar to the average, is forward-propagated through the \textit{Seed Network}. Figure~\ref{sub:A}, is the feature map with no priming.  The values are fairly equally spread out. Figure~\ref{sub:B} indicates the outputs that are eliminated when the $\pm$ 1$\sigma$ threshold is applied.  Black indicates the outputs that are not connected. Figure~\ref{sub:C} is the feature map after 18 priming cycles have been executed.  The values are more concentrated near mid gray with some outputs that are very dark or very light.  Figure~\ref{sub:D} indicates that outputs that are eliminated when the $\pm$ 1$\sigma$ threshold is applied.  Black indicates the outputs that are not connected.  Here we see more defined groups of outputs that are not connected. There are three bands of 9 outputs entirely black.  Because the outputs all belong to the same kernel, the entire kernel can be eliminated.
\end{comment}
%\fi
\subsection{Growing the Most Accurate Architecture}\label{subsec:PrimeSweep}
\vspace{-2mm}
%\subsubsection{Purpose}
We sweep \textit{priming cycles} with scaling factor of 1.0 to find the number of \textit{priming cycles} needed to grow our most accurate network. 
%We are now ready to grow our most accurate network. Using a scaling factor of 1.0, we sweep the \textit{priming cycles}. Our goal is to find the number of \textit{priming cycles} it takes to produce the best network. 
%\subsubsection{Procedure}
One at a time, we prepare a \textit{\textit{Seed Network}} by varying the  \textit{priming cycles} from 1 to 25. We add the extreme perceptrons and connect \textit{source layer} outputs \begin{math} \pm \end{math}1.0 $\sigma$ from the average to the \textit{destination layer}. We train the entire network until one of the stopping criteria are met.
%\subsubsection{Analysis}
In Table \ref{tbl:FullNetworkSigma}, we observe that the number of critical connections reduces as we increase the number of \textit{priming cycles}.
This effect is not seen during the earlier \textit{priming cycle} sweep experiment presented with Table \ref{tbl:SeedNetworkPriming} where the \textit{priming cycles} were held to 18. As we sweep the number of priming cycles, the testing inference accuracy varies between 98.28\% and 98.80\% with a peak validation accuracy on the eleventh priming cycle. 

Based on the sweeping based experiments presented above, the configuration of the final ANG generated architecture is shown in Table \ref{tbl:NeurogenesisNetwork}. The fully connected layer in this table is a direct result of the best scaling factor of 1$\sigma$ found in Table~\ref{tbl:SeedNetworkSigma} and the number of priming cycles of 11 found in Table~\ref{tbl:FullNetworkSigma}.

%mixter
%Once the best accuracy is achieved after the stopping criteria has been met, we are at the point where the above mentioned pruning methods start.  By sweeping the various parameters, we have grown a small, well trained neural network that is ready for pruning.

\begin{table}[t]
\small
\caption{Most Accurate Network Grown using ANG}\label{tbl:NeurogenesisNetwork}
\centering
\begin{tabular}{|c|c|c|c|r|r|}
\hline
%\rowcolor[HTML]{EFEFEF} 
Layer      & Filter & Kernel & Stride & Perceptron & Weight \\ \hline
2D Conv & 6 & 7 & 2 & 864 & 300 \\ \hline
3D Conv & 50 & 7 & 4 & 450 & 14,750 \\ \hline
Full &  &  &  & 20 & 5,951 \\ \hline
Classifier &  &  &  & 10 & 210 \\ \hline
\multicolumn{4}{|r|}{Totals}            & 1,344        & 21,211    \\ \hline
\end{tabular}
\vspace{-6mm}
\end{table}

\begin{table}[t]
\small
\caption{Network Size Relative to Grown ANG Network}\label{tbl:Relative}
\centering
%\small
\begin{tabular}{|c|c|c|c|c|}
\hline
%\rowcolor[HTML]{EFEFEF} 
\begin{tabular}[c]{@{}c@{}}Neural\\ Network\end{tabular} &
\begin{tabular}[c]{@{}c@{}}Starting\\ Weight\end{tabular} &
\begin{tabular}[c]{@{}c@{}}Final\\ Weight\end{tabular} & \begin{tabular}[c]{@{}c@{}}Relative\\ Size \%\end{tabular} & \begin{tabular}[c]{@{}c@{}}Test\\ Accy\end{tabular} \\ \hline
Babaeizadeh et al.\cite{Babaeizadeh:Illinois} & 606K & 13.6K & -35.9 & 99.1 \\ \hline
Tu et al.\cite{Tu:ASU} & 606K & 32K & 50.9 & 98.4 \\ \hline
Han et al.\cite{Han:Stanford}& 431K & 36K & 69.7 & 99.2 \\ \hline
Blundel et al.\cite{Blundell:Google}& 2.4M & 48K & 126 & 98.6 \\ \hline
Srinivas et al.\cite{Srinivas:IIS}& 431K & 71K & 235 & 98.4 \\ %\hline
\hlinewd{1pt}
ANG Grown & 21.2K & 21.2K & 0.0 & 98.8 \\ \hline
%Pruned ANG & 18.2k & 18.2k & 0.0 & 98.8 \\ \hline
\end{tabular}
\vspace{-4mm}
\end{table}

\vspace{-3mm}
\section{Artificial Neurogenesis Analysis}
\vspace{-2mm}
\label{sec:NeurogenesisAnalysis}
The goal of ANG is to grow neural networks that are small enough to fit into low SWaP embedded hardware while still performing as well as full sized networks.  
ANG starts with a \textit{Seed Network} that has 19,560 weights (Table~\ref{tbl:SeedNetwork}) and then
%, which is smaller than all but one of the pruned networks.  
as shown in Table \ref{tbl:NeurogenesisNetwork}, the \textit{Seed Network} is grown to a size of 21,211 weights. 
%The ANG grown network is then reduced using weight magnitude pruning to a final size of 18,167 weights (Table~\ref{tbl:Reduction}) while maintaining a test inference accuracy of 98.80\%. Table~\ref{tbl:Reduction} also shows the degree of pruning that occur in each layer. The largest reduction occurs in the Fully Connected layer, which is where ANG takes place. The next largest reduction is the result of pruning the classifying layer. Because the convolution layers are more sensitive to pruning due to sharing of weights, these layers are relatively untouched.

In Table \ref{tbl:Relative}, we present our final comparison in terms of number of weights (Starting and Final), Relative Size in terms of the percentage of change in number of weights relative to the grown ANG network, and Test Accuracy. Based on the starting weight sizes, none of the pruning based methods are as suitable as our ANG Grown method for execution in low SWaP hardware. In terms of the final number of weights, the grown ANG network is smaller than all but one network \cite{Babaeizadeh:Illinois}. In terms of accuracy, the grown ANG network is less accurate than only two networks (~\cite{Babaeizadeh:Illinois} and \cite{Han:Stanford}). In overall, we believe that unlike other methods, the ANG offers ability to train and infer completely on-chip with a slight trade-off in accuracy. %The pruned ANG network is 16.8\% smaller than the ANG grown network and 70.30\% smaller than our baseline network. The ANG grown network and pruned ANG networks are slightly more accurate than the baseline neural network.
Another advantage of our method is that it offers faster training and inference time due to much smaller starting network size.

%The goal of ANG is to grow neural networks that are small enough to fit into low SWaP embedded hardware while still performing at least as well as full sized networks. 
Artificial Neurogenesis (ANG) is the opposite of pruning.  ANG \textit{grows} a network from a small seed using only the resources required to achieve a size of 21,211 weights. Pruning methods on the other hand demand for much more resources to store and process a full network of weights ranging from 400,000 \cite{Srinivas:IIS} to 2.4 million \cite{Blundell:Google} before training and reducing to scale of 13,579 weights. 

The proposed ANG grows a network whose testing accuracy (98.80\%) is comparable to the best pruned accuracy (99.05\%), only differing by 0.25\%.
We finally evaluate the efficiency of the ANG method based on its ability to find connections that are more critical than the connections revealed by weight magnitude pruning. The contribution of the proposed algorithm can only be validated if a normal network with same structure cannot reach the same performance. To determine this, we will build a network whose architecture is identical to our ANG network, Tables \ref{tbl:NeurogenesisNetwork} and Table \ref{tbl:20FC}, train the network until the stopping criteria are met and compare the results. As shown in Table \ref{tbl:20FCPrune}, the 20 FC network testing accuracy never reaches the accuracy of 98.80\% achieved by our ANG grown network. If we prune the network to nearly the same size as our grown network (bold row, 11.80\% removed), the Test Accuracy is significantly less.  More importantly, even with the 20 FC network not pruned (0\% Removed), the achieved testing accuracy is less than the ANG grown network. If we compare the two networks by their connection counts, the ANG grown network is significantly more accurate. This gives us confidence that the algorithm is performing well and is choosing critical connections. 

%Given the ten classes of MNIST and the two extreme perceptrons added for each class, the grown layer has 20 perceptrons.  We create a network similar to our baseline except the fully connected (FC) layer has 20 perceptrons instead of 100.  The network architecture (20 FC), is shown in Table \ref{tbl:20FC}. After we create a 20 FC network, we train until the stopping criteria are met.  Once the network is trained, The new network, 20 FC, is trained until the stopping criteria are met. } We then adjust the pruning threshold to achieve the percent removed for each row in Table \ref{tbl:20FCPrune}.  After the network is pruned, we execute a testing inference and note the accuracy.

\begin{table}
\caption{20 Perceptron Fully Connected Network (20 FC)}\label{tbl:20FC}
\centering
\small
\begin{tabular}{|c|c|c|c|r|r|}
\hline
%\rowcolor[HTML]{EFEFEF} 
Layer      & Filters & Kernel & Stride & Perceptrons & Weights \\ \hline
2D Conv    & 6       & 7      & 2      & 864         & 300     \\ \hline
3D Conv    & 50      & 7      & 4      & 450         & 14,750   \\ \hline
Full       &         &        &        & 20          & 9,020    \\ \hline
Classifier &         &        &        & 10          & 210     \\ \hline
\multicolumn{4}{|r|}{Totals}            & 1,344        & 24,280    \\ \hline
\end{tabular}
\vspace{-4mm}
\end{table}

\begin{table}[!t]
\caption{Pruned 20 Perceptron Fully Connected Network}\label{tbl:20FCPrune}
\centering
\small
\begin{tabular}{|c|c|c|c|}
\hline
%\rowcolor[HTML]{EFEFEF} 
\% Removed & Connections & Test Accuracy & Error \\ \hline
0\% & 24,280 & 98.50\% & 1.50\% \\ \hline
1.77\% & 23,850 & 98.50\% & 1.50\% \\ \hline
5.79\% & 22,874 & 98.49\% & 1.51\% \\ \hline
6.18\% & 22,779 & 98.48\% & 1.52\% \\ \hline
7.39\% & 22,486 & 98.46\% & 1.54\% \\ \hline
8.54\% & 22,207 & 98.47\% & 1.53\% \\ \hline
\textbf{11.80\%} & \textbf{21416} & \textbf{98.39\%} & \textbf{1.61\%} \\ \hline
15.18\% & 20,595 & 98.37\% & 1.63\% \\ \hline
18.24\% & 19,852 & 98.26\% & 1.74\% \\ \hline
20.90\% & 19,205 & 98.01\% & 1.99\% \\ \hline
23.28\% & 18,627 & 97.59\% & 2.41\% \\ \hline
25.25\% & 18,150 & 95.97\% & 4.03\% \\ \hline
26.98\% & 17,729 & 93.82\% & 6.18\% \\ \hline
28.46\% & 17,369 & 91.85\% & 8.15\% \\ \hline
36.00\% & 15,538 & 61.10\% & 38.90\% \\ \hline
37.61\% & 15,148 & 33.29\% & 66.71\% \\ \hline
\end{tabular}
\vspace{-6mm}
\end{table}

\vspace{-2mm}
\section{Conclusions and Future Work}
\vspace{-4mm}
\label{sec:Conclusion}
%In this study, our goal is to create neural networks that can fit into low SWaP embedded hardware. 
%This will allow edge devices to learn and adapt autonomously. Because the devices we are targeting have low SWaP requirements, they cannot fit state-of-the-art large neural networks. 
We propose a network growing method as an alternative to traditional pruning based approaches so that we avoid the need for offline training and make training a feasible process under low SWaP requirements. Our algorithm, Artificial Neurogenesis (ANG), grows neural networks from small \textit{Seed Networks} by identifying critical outputs in the \textit{source layer} and connecting them to \textit{\textit{destination layer}}. Once a network has been grown and trained, ANG then applies pruning as a final step to further reduce the size of the neural network. Working with the MNIST data set, we applied ANG to grow a neural network that achieves 98.8\% inference accuracy on the test data set. 

One of the main conclusions that can be reached from this research is that the training data holds information that can be used to determine network architecture prior to training.  In this instance, we only targeted the fully connected layer but, ANG can be applied to the connections to the classifying layer. This will be investigated in future research along with analyzing the convolution layers to find critical connections.  

The neural network we grew only required 21.2K weights to support this accuracy. To the best of our knowledge, no other method in the literature uses input data to determine an architecture that can be implemented on low SWaP hardware. Because we were able to prune the grown layer, it is reasonable to assume we have not found the optimal critical connections. Therefore, future research will be directed at further analyzing the \textit{source layer} outputs. One change would be to use the average image data as the target for analysis instead of an actual image close to the average image data. It is our belief that neural networks can be grown to near optimal architectures using input data as a guide. ANG has proven to be a step in that direction.    
%\vspace{-4mm}

\section*{Acknowledgment} \label{sec:acknowledgement}
\vspace{-4mm}
 \footnotesize{
 This work is partly supported by National Science Foundation research project CNS-1624668 and Raytheon Missile Systems (RMS) under the contract 2017-UNI-0008. The content is solely the responsibility of the authors and does not necessarily represent the official views of RMS.
 }

\bibliographystyle{spphys}
\bibliography{references}

\begin{comment}
 
\end{comment}
\end{document}